\documentclass[letterpaper, 10 pt, conference]{ieeeconf}
\IEEEoverridecommandlockouts
\usepackage{cite}
\usepackage{amsmath,amssymb,amsfonts}
\usepackage{algorithmic}
\usepackage{graphicx}
\usepackage{textcomp}
\usepackage{xcolor}
\usepackage{caption}
\usepackage{booktabs}
\usepackage{algorithm}
\usepackage{multirow}
\usepackage{tabularx}
\usepackage{dsfont}

\def\BibTeX{{\rm B\kern-.05em{\sc i\kern-.025em b}\kern-.08em
    T\kern-.1667em\lower.7ex\hbox{E}\kern-.125emX}}
\begin{document}

\title{TacHOI: Contact-Aware Object Reconstruction and Tracking \\for Egocentric Videos\\

}

\author{Jackson Codd, Haoyu Li, Xianyi Cheng}

\maketitle

\begin{abstract}
Egocentric videos offer a scalable source of human demonstrations for robot learning. However, recovering metric object geometry and motion from monocular video is hindered by scale–depth ambiguity and hand-induced occlusion, while image-derived contact and shape priors alone cannot uniquely determine metric scale. We introduce TacHOI, a training-free framework that integrates tactile contact with monocular vision for joint hand–object reconstruction and tracking, without requiring object templates or depth sensors. Starting from pretrained visual estimates, TacHOI uses tactile contact to constrain object depth relative to the estimated hand, while spatially separated contacts on a metrically calibrated hand provide a physical ruler for object extent. It incorporates these constraints into a joint optimization of hand poses, object trajectories, and anisotropic object scale alongside visual alignment. Across 100 ARCTIC and HO3D sequences with synthesized tactile observations, TacHOI achieves the lowest median ADD-S* among the evaluated baselines, reducing error from 15.6 to 3.6 cm on ARCTIC and achieving 2.8 cm on HO3D. Additional experiments on OpenTouch demonstrate its applicability to measured tactile observations.

\end{abstract}

% \begin{IEEEkeywords}
% Object Tracking, 3D Reconstruction, Tactile Sensing, Egocentric Videos
% \end{IEEEkeywords}

\section{Introduction}

Humans effortlessly manipulate objects in daily life, from pouring a drink to assembling a tool. The growing adoption of smart glasses~\cite{projectaria} offers new opportunities to capture these everyday interactions at scale, turning daily activities into a rich source of demonstrations for robots to learn manipulation skills~\cite{egomimic}. However, a key challenge of learning manipulation skills from egocentric videos is the ambiguity in estimating precise hand-object trajectories, especially for monocular observation. Hand-induced occlusion and self-occlusion further complicate accurate reconstruction and tracking~\cite{magichoi}.

A fundamental challenge in monocular reconstruction is scale ambiguity: an object scaled by a factor of $k$ and placed $k$ times farther from the camera produces the same image projection. Even accurate masks and 2D tracks therefore cannot uniquely determine its metric scale and depth. When demonstrations are retargeted to a robot, scale errors can lead to misplaced grasps and incorrect motion trajectories. Existing methods rely on additional information, including known instance-level 3D meshes~\cite{foundationpose}, depth measurements~\cite{bundlesdf}, multiple camera views~\cite{neuraldome}, object scans~\cite{imhoi}, or learned shape and grasp priors~\cite{ghop}. These requirements constrain their applicability to in-the-wild human demonstrations. Meanwhile, growing efforts to capture such demonstrations with tactile gloves~\cite{opentouch,feeltheforce,osmo} offer a complementary source of geometric constraints. Unlike contact inferred from vision, measured tactile contact provides direct physical evidence of hand--object contact, which can constrain reconstruction when paired with calibrated hand geometry.

\begin{figure}[t]
    \centering
    \includegraphics[width=\linewidth]{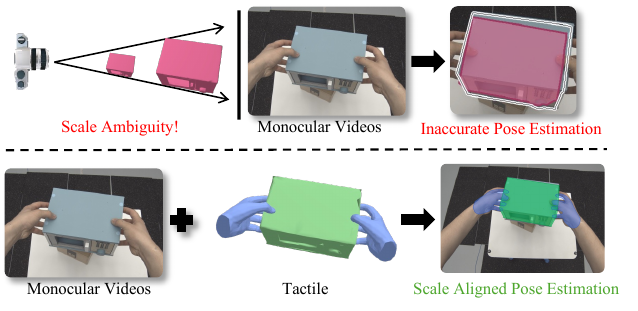}
    \caption{Monocular observations allow for multiple object depths and scales
    with similar image projections. TacHOI uses tactile contact with a metrically
    calibrated hand to constrain the object's camera-space depth and anisotropic
    scale.}
    \label{fig:teaser}
    \vspace{-4mm}
\end{figure}

To address these challenges, we introduce \emph{TacHOI}, a training-free hand-object reconstruction framework that leverages tactile contact as an additional geometric constraint for hand-object reconstruction and tracking. Unlike contact inferred from vision, measured tactile contact provides physical evidence that is independent of the object's image projection. \emph{TacHOI} uses this evidence in two ways. First, tactile contact pulls the reconstructed hand and object together. Second, \emph{TacHOI} uses the calibrated hand as a physical ruler. Spatially separated contacts provide a caliper-like measurement of the object extent between the contacting parts of the hand. This gives the optimizer a metric constraint on object scale without requiring a known object template or depth sensor. Based on the constraints, the framework jointly optimizes the object's per-frame pose, anisotropic scale, and the hand pose. In summary, our main contributions are:
\begin{itemize}
\item We present a framework that requires no task-specific training for jointly refining object reconstruction, metric scale, 6-DoF pose, and hand pose from monocular egocentric video and tactile contact. \item We show that spatially separated tactile contacts on a calibrated hand act as a metric constraint, resolving the monocular scale-depth ambiguity without object templates or depth sensing. \item We evaluate TacHOI on 100 ARCTIC and HO3D clips under a controlled tactile-synthesis protocol, showing consistent improvements in object scale and camera-space localization over vision-based initialization.
\end{itemize}

\section{Related Work}

\subsection{Model-Free Object Pose Estimation}

Most object-pose estimators assume a CAD model at test time. Model-free methods instead construct object representations from reference observations or learned geometric priors. FoundationPose~\cite{foundationpose} at its core is a CAD-based method, but it supports model-free pose estimation from reference images. Any6D~\cite{any6d} reconstructs an object representation from a single RGB-D observation and jointly estimates metric size and 6-DoF pose. Although effective on unseen objects, these approaches still rely on reference observations or depth and generally do not exploit physical constraints from hand-object interaction. Recent methods instead reconstruct object motion directly from monocular interaction videos. Do-as-I-Do~\cite{do_as_i_do}, MagicHOI~\cite{magichoi}, Follow My HOLD~\cite{followmyhold}, Grasp in Gaussians~\cite{graspingaussians}, AGILE~\cite{agile}, and CHOIR~\cite{choir} combine hand-object interaction cues with learned geometric, generative, or contact priors to improve reconstruction and tracking. These methods demonstrate the value of interaction context, but their contact constraints are inferred from images and therefore remain sensitive to occlusion and errors in monocular depth, scale, and hand pose. TacHOI instead uses measured pressure to provide direct evidence of when and where contact occurs.

\subsection{Object Reconstruction}

Classical geometry-based methods such as GSLAM~\cite{gslam} recover scene structure through multi-view consistency and camera motion estimation. Reconstruction of handheld objects is more challenging because both the camera and the object may move. Hands also cause severe occlusion, and monocular geometry remains ambiguous in scale. Sequence-level neural reconstruction methods address these challenges by jointly modeling hands and objects. HOLD~\cite{hold} uses compositional implicit fields for category-agnostic reconstruction, while G-HOP~\cite{ghop} and MagicHOI~\cite{magichoi} introduce learned generative or novel-view priors to recover geometry under limited observations. Such priors enable plausible completion of occluded regions but do not necessarily recover the metric configuration of the specific observed object. Feed-forward models provide a faster alternative. SAM 3D Objects~\cite{sam3d} reconstructs complete objects from masked images, while ForeHOI~\cite{forehoi} incorporates video evidence and hand-object-specific completion. These models provide useful geometric priors, but image plausibility and silhouette consistency alone do not determine metric scale or depth. Additionally, pretrained models are limited by the scope of their training data. For naturally ambiguous objects, like bananas, SAM 3D cannot reconstruct with the correct aspect ratios. TacHOI complements pretrained reconstruction with measured contact, introducing a new information source to ground metric pose estimation and reconstruction.

\subsection{Visuotactile Hand--Object Estimation}

Contact has previously been used as a geometric prior without direct tactile sensing. Contact Potential Fields (CPF)~\cite{cpf} optimize predicted hand-object contacts toward physically plausible configurations, but the contact itself is inferred visually and is therefore unreliable under severe occlusion. Physical tactile sensing can provide stronger constraints. Tac2Pose~\cite{tac2pose} estimates object pose from optical tactile observations, while NeuralFeels~\cite{neuralfeels} combines RGB-D, tactile sensing, and robot proprioception to jointly estimate object shape and pose. The ViTaM family~\cite{vitam,vitamd} similarly fuses vision with distributed tactile sensing for reconstruction during deformable or forceful interactions. Mack et al. ~\cite{mack} further show that low-resolution binary contact improves RGB-D object-pose estimation within a factor-graph formulation. These approaches demonstrate the value of tactile information but typically assume known object geometry, depth sensing, calibrated robot kinematics, specialized tactile hardware, or learned object-specific models. TacHOI targets a complementary setting: monocular reconstruction of an unknown-category object manipulated by an instrumented human hand. Rather than reconstructing local surface geometry from tactile images, TacHOI converts low-resolution pressure measurements into contact constraints between the estimated hand and object, providing metric information for object depth, scale, and pose without a CAD template, depth sensor, or calibrated robot proprioception.
\section{Method}
\label{sec:method}

\begin{figure*}[t]
    \centering
    \includegraphics[width=.95\textwidth]{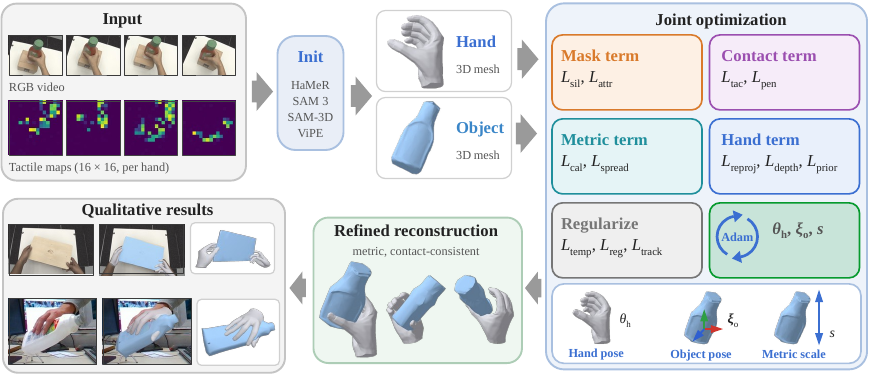}
    \caption{Overview of TacHOI. Off-the-shelf models initialize the hands, object, camera trajectory, and amodal masks from monocular RGB. TacHOI jointly refines hand and object trajectories and anisotropic object scale. Image-space terms constrain projected alignment, while measured contact constrains relative depth and supplies metric object-extent measurements.}
    \label{fig:method-overview}
\end{figure*}

\subsection{Overview and Initialization}
\label{sec:initialization}

TacHOI jointly refines hand pose, object pose, and anisotropic object scale using complementary visual and tactile cues. Silhouettes and sparse tracks align the object with the image, tactile contact anchors it to the hand, and the metrically calibrated hand constrains object extent through caliper and contact-spread measurements.

The OpenTouch~\cite{opentouch} dataset provides a $16\times16$ pressure map. For benchmarks without tactile measurements, we synthesize maps from ground-truth hand and object poses by assigning palmar-facing MANO vertices to their nearest taxels and measuring their distance to the object mesh. Pressure activates within 3\,mm of contact and increases linearly to 1 as the distance approaches zero, approximating sensor compliance while preserving pressure magnitude as a continuous per-taxel weight. A YOLO-based detector~\cite{yolo} localizes each hand, and HaMeR~\cite{hamer} initializes its MANO~\cite{mano} parameters and 2-D joints. Metric hand scale is obtained from Rokoko measurements on OpenTouch~\cite{opentouch} and ground-truth MANO bone lengths on ARCTIC~\cite{arctic} and HO3D~\cite{ho3d}.

For the object, Qwen3-VL~\cite{qwen3vl} and SAM 3~\cite{sam3} produce a modal mask, which Diffusion-VAS~\cite{diffusionvas} completes to an amodal mask $M_t$. SAM 3D Objects~\cite{sam3d} reconstructs a canonical mesh $\mathcal V_0=\{\mathbf v_j\}_{j=1}^{J}$, Do-As-I-Do~\cite{do_as_i_do} initializes its camera-space pose and trajectory, and CoTracker3~\cite{cotracker3} supplies sparse 2-D tracks for when the object is off-screen. ViPE~\cite{vipe} provides camera-to-world transforms.

We convert HaMeR's weak-perspective translation into a metric wrist-depth estimate $\hat d_t^h$ using the focal length and calibrated hand size. For contact confidence $g_t^h>0.3$, these estimates provide a sparse object-depth proxy. We use it to compute one robust metric conversion for the complete Do-As-I-Do trajectory:
\begin{equation}
\begin{aligned}
d_t^{\mathrm{obj}}
&=
\frac{\sum_h \mathbf{1}[g_t^h>0.3]\,\hat d_t^h}
     {\sum_h \mathbf{1}[g_t^h>0.3]},\\
\kappa_{\mathrm{metric}}
&=
\underset{t\in\mathcal T_d}{\operatorname{median}}
\frac{d_t^{\mathrm{obj}}}
     {[\widetilde{\mathbf t}_t^{\mathrm{trk}}]_z},
\qquad
s_0=\kappa_{\mathrm{metric}}\widetilde{s}_{t_a}^{\mathrm{trk}} .
\end{aligned}
\label{eq:metric-conversion}
\end{equation}
Here, $\mathcal T_d$ contains valid contact-depth frames and $t_a$ is the SAM 3D anchor frame. This conversion initializes metric object scale but does not impose the wrist-depth estimate independently in every frame.

\subsection{Hand and Object Parameterization}
\label{sec:parameterization}

For each frame, we optimize object rotation $\mathbf r_t^o\in\mathbb R^6$, translation $\mathbf t_t^o\in\mathbb R^3$, and a global log-scale correction $\boldsymbol\alpha\in\mathbb R^3$ along the canonical principal axes. Anisotropic scaling allows refinement of aspect ratios that are weakly observed from the single SAM 3D anchor view. With mesh centroid $\mathbf c_0$ and principal-axis basis $U_{\mathrm{PCA}}$, the scaled and transformed vertices are
\begin{align}
\mathcal A_{\boldsymbol\alpha}(\mathbf v)
&=
\mathbf c_0+
U_{\mathrm{PCA}}^{\!\top}
\operatorname{diag}(e^{\boldsymbol\alpha})
U_{\mathrm{PCA}}(\mathbf v-\mathbf c_0),
\nonumber\\
\mathbf x_{t,j}^{o}
&=
s_0R(\mathbf r_t^o)
\mathcal A_{\boldsymbol\alpha}(\mathbf v_j)
+\mathbf t_t^o .
\label{eq:object-transform}
\end{align}
Object parameters are optimized in world coordinates. Then, fixed ViPE-estimated camera extrinsics transform them into the camera frame for projection.

For each hand $h$, we optimize one shared MANO shape $\boldsymbol\beta^h\in\mathbb R^{10}$ and per-frame articulation $\boldsymbol\theta_t^h\in\mathbb R^{45}$, wrist rotation $\mathbf r_t^h\in\mathbb R^6$, and translation $\mathbf t_t^h\in\mathbb R^3$. We denote the collective hand parameterization by $\Psi^h$.
\subsection{Joint Refinement}
\label{sec:optimization}

We optimize both hands and the object jointly. At each iteration, we sample a shared frame batch $\mathcal B\subseteq\{1,\ldots,N\}$ with $|\mathcal B|=12$. For any validity set $\mathcal T_X$, let $\mathcal B_X=\mathcal B\cap\mathcal T_X$. Unless otherwise stated, all refinement losses are evaluated over their corresponding valid minibatch. For each EMA-normalized image term with raw minibatch loss $\ell_X$, we use \begin{equation}
\mathcal N(\ell_X)=
\frac{\ell_X}
{\operatorname{sg}(\operatorname{EMA}[\ell_X])+\epsilon}. 
\label{eq:ema-normalization}
\end{equation} For the image-space terms, lowercase $\ell_X$ denotes the raw loss, and uppercase $\mathcal L_X$ denotes the value normalized by a stop-gradient exponential moving average.
\begin{equation}
\begin{aligned}
\operatorname*{minimize}\quad
 \mathcal{L}_{\mathrm{obj}}
  + \sum_{h=1}^{H}\mathcal{L}_{\mathrm{hand}}^h \\
\text{over}\quad
 \{\boldsymbol{\xi}_t^o\}_{t=1}^{N},
  \ \boldsymbol{\alpha},
  \ \{\Psi^h\}_{h=1}^{H}
\end{aligned}
\label{eq:joint-objective}
\end{equation}
The object objective combines image, tactile, and prior terms. We selected the loss coefficients empirically by evaluating a range of values on representative sequences and iteratively adjusting them based on reconstruction accuracy and optimization stability. The complete object objective is: 
\begin{align}
\begin{split}
\mathcal{L}_{\mathrm{obj}}={}
1.5\mathcal{L}_{\mathrm{sil}}
+\mathcal{L}_{\mathrm{attr}}
+10\mathcal{L}_{\mathrm{track}}
\\
+\mathcal{L}_{\mathrm{tac}}
+3\mathcal{L}_{\mathrm{cal}}
+\mathcal{L}_{\mathrm{reg}}
\label{eq:object-objective}
\end{split}
\end{align}
All contact locations, nearest surface points, and penetration gates are recomputed from the current hand and object parameters at every iteration. The image terms $\mathcal L_{\mathrm{sil}}$, $\mathcal L_{\mathrm{attr}}$, $\mathcal L_{\mathrm{track}}$, and $\mathcal L_{\mathrm{reproj}}^h$ are normalized by exponential moving averages. Consequently, their coefficients specify target contributions rather than raw-loss weights.

\subsection{Object Measurement Terms}
\label{sec:object-terms}

Let $\pi:\mathbb R^3\rightarrow\mathbb R^2$ denote perspective projection with known camera intrinsics, let $\eta=H_I+W_I$ normalize pixel distances, and let $\operatorname{sg}(\cdot)$ denote stop-gradient.

\subsubsection{Silhouette Alignment and Mask Attraction}
\label{sec:image-terms}

Let $S_t(p)\in[0,1]$ be the differentiable rendered silhouette and $M_t(p)\in\{0,1\}$ the amodal mask. Frames without a valid pose, camera, or object mask are excluded from the object objective altogether. To down-weight truncated masks, we define a border weight:
\begin{equation}
w_t^{\mathrm{border}}=1-\frac{B_t}{C_t}
\label{eq:border-weight}
\end{equation}
Here $B_t$ is the length of the mask contour coincident with the image border and $C_t$ is its total contour length. For valid-mask frames $\mathcal B_M$,
\begin{align}
\ell_{\mathrm{sil}}&=
\frac{1}{|\mathcal B_M|}\sum_{t\in\mathcal B_M}
w_t^{\mathrm{border}}\frac{1}{H_IW_I}
\sum_p\bigl(S_t(p)-M_t(p)\bigr)^2
\label{eq:silhouette}
\end{align}
Silhouette overlap provides a strong local signal for projected extent, but its gradients can vanish when the initial projection and mask do not overlap. We therefore sample pixels from the target mask and attract them to projected mesh vertices. The per-pixel cost is the distance to the nearest vertex, discounted by that vertex's visibility confidence.
\begin{equation}
d_t(p)=
\frac{1}{\eta}\min_j
\left(\|p-\pi(\mathbf x_{t,j}^o)\|_2
-\tau_v\log w^{\mathrm{vis}}_{t,j}\right)
\label{eq:attraction-cost}
\end{equation}
Here $w^{\mathrm{vis}}_{t,j}$ is a detached confidence that discounts self-occluded vertices and vertices near the image boundary. The attraction term complements the silhouette term by computing the pixel distance rather than the pixel overlap of the mask and object projection. Therefore, the attraction term remains active even when the object render does not overlap the mask. Pixels are importance-sampled toward the part of the mask the current silhouette does not yet cover (2000 samples per frame), so the term concentrates on the region the silhouette has not reached.
\begin{equation}
\begin{split}
\ell_{\mathrm{attr}}&=
\frac{1}{|\mathcal B_M|}\sum_{t\in\mathcal B_M}
\mathbb E_{p\sim q_t}\big[d_t(p)\big]
\label{eq:attraction}
\end{split}
\end{equation}
The attraction term guides the object toward the observed mask when the initial overlap is poor, after which the silhouette term provides dense alignment.

\subsubsection{Sparse Point Tracks}
\label{sec:track-term}

Silhouettes don't provide surface correspondence and become ambiguous when truncated. Therefore, we use tracks only when the object leaves the image. Tracks are lifted from up to six frames to improve surface coverage as the object rotates. The lifting frames are the SAM 3D anchor and the frame with the largest fully visible mask in each of five clip segments. We compare each lifted point $\mathbf q_m$ with its observed projection $\widehat{\mathbf u}_{t,m}$:
\begin{equation}
\begin{split}
e_{t,m}&=\frac{\|\pi(\mathcal G_t^o(\mathbf q_m))-
\widehat{\mathbf u}_{t,m}\|_2}{\eta},
\\
\rho_c(e)&=\frac{e^2}{1+e^2/c^2}.
\label{eq:robust-kernel}
\end{split}
\end{equation}
For minibatch $\mathcal B_{\mathrm{trk}}$, the gated track loss is defined with
$c_{\mathrm{trk}}=30\,\mathrm{px}/\eta$ as
\begin{align}
\bar\gamma_{\mathcal B_{\mathrm{trk}}}&=\frac{1}{|\mathcal B_{\mathrm{trk}}|}
\sum_{t\in\mathcal B_{\mathrm{trk}}}(1-w_t^{\mathrm{border}})^2\\
\ell_{\mathrm{track}}&=
\frac{\sum_{t\in\mathcal B_{\mathrm{trk}},m}\gamma_t v_{t,m}\,
\rho_{c_{\mathrm{trk}}}(e_{t,m})}
{\sum_{t\in\mathcal B_{\mathrm{trk}},m}\gamma_t v_{t,m}+\epsilon}
\label{eq:track-loss}
\end{align}
Here, $v_{t,m}$ denotes track visibility. The inner gate emphasizes truncated frames, while the outer factor $\bar\gamma_{\mathcal B_{\mathrm{trk}}}$ preserves the gate's effect after EMA normalization. This prevents tracks, which inherit stage-1 lifting error, from overriding reliable silhouettes on fully visible frames while retaining their correspondence signal under truncation.

\subsubsection{Initialization and Static-Scene Priors}
\label{sec:object-priors}

Let $u_t$ be the fraction of the initial rendering not covered by the mask and
$a_t=[1+(u_t/0.15)^2]^{-1}$. We regularize well-initialized frames toward
their initial poses. Within each confidently static interval, all frames share
one rotation variable, and adjacent translations are encouraged to remain
constant. The combined regularizer is
\begin{align}
\mathcal L_{\mathrm{reg}}={}&
\frac{1}{|\mathcal B|}\sum_{t\in\mathcal B}a_t
\left(
20\|\mathbf t_t^o-\mathbf t_t^{o,0}\|_2^2
+5\|\mathbf r_t^o-\mathbf r_t^{o,0}\|_2^2
\right)
\nonumber\\
&+\frac{3\times10^4}{|\mathcal S_{\mathrm{static}}|}
\sum_{(t,t+1)\in\mathcal S_{\mathrm{static}}}
\|\mathbf t_{t+1}^o-\mathbf t_t^o\|_2^2 .
\label{eq:object-regularization}
\end{align}

\subsection{Tactile Measurement Terms}
\label{sec:tactile-terms}

\subsubsection{Correspondence-Free Tactile Contact}
\label{sec:tactile-contact}

We compute a contact confidence $g_t^h\in[0,1]$ from the pressure signal and consider contact active when $g_t^h>0.3$. Let $\mathcal C_{\mathrm{DGN}}^h$ contain the 84 contact-candidate entries provided by the official DexGraspNet~\cite{dexgraspnet} MANO implementation. Rather than assigning each tactile observation to a specific vertex, we compute a soft minimum over the entire candidate set:
\begin{equation}
\begin{split}
d_t^{h,\min}&=-\tau_c\log
\sum_{\mathbf c\in\mathcal C_{\mathrm{DGN}}^h}
\exp\!\left[
-\frac{d(\mathbf c,\mathcal G_t^o(\mathcal V_0))}{\tau_c}
\right],
\\
\tau_c&=5\,\mathrm{mm}.
\label{eq:contact-softmin}
\end{split}
\end{equation}

\begin{figure*}[!t]
\centering
\includegraphics[
    width=0.8\textwidth,
    keepaspectratio
]{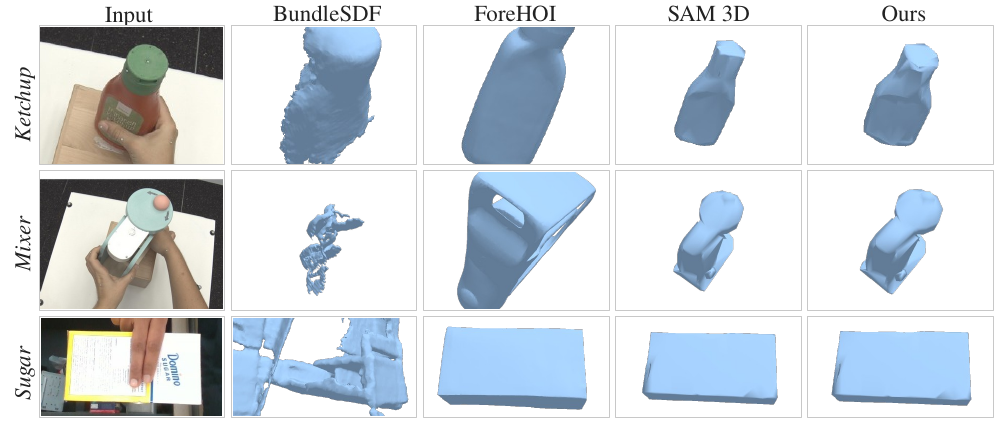}
\caption{Qualitative reconstruction comparison. Each benchmark row shows a common canonical view of the reconstruction baselines, SAM 3D initialization, TacHOI, and ground truth. The results are very similar to SAM 3D since we do not refine object shape. Instead, only the anisotropic axes are modified, allowing us to optimize the reconstruction without distorting the object.}
\label{fig:recon-comparison}
\end{figure*}

This correspondence-free constraint resolves along-ray depth ambiguity by requiring the object surface to reach the hand somewhere. Finger identity is used only by the scale constraints below.

We define $\widetilde\omega_t^{\mathrm{tac},h}$ from the pressure confidence, retaining plausible contacts when visual evidence is weak and down-weighting initial hand--object offsets that are primarily lateral to the camera ray. The filter is fixed before refinement to prevent lateral hand-tracking errors from pulling the object away from image support. The tactile loss is
\begin{equation}
\begin{split}
\mathcal L_{\mathrm{tac}}&=10^3
\frac{\sum_{t\in\mathcal B_{\mathrm{tac}}}\sum_h\widetilde\omega_t^{\mathrm{tac},h}
\rho_{c_{\mathrm{tac}}}(d_t^{h,\min})}
{\sum_{t\in\mathcal B_{\mathrm{tac}}}\sum_h\widetilde\omega_t^{\mathrm{tac},h}+\epsilon},
\\
c_{\mathrm{tac}}&=10\,\mathrm{cm}.
\label{eq:tactile-loss}
\end{split}
\end{equation}
\subsubsection{Hand-as-Caliper Scale Constraint}
\label{sec:caliper}

The calibrated hand provides a metric constraint. We use the taxel-to-finger mapping to construct contact patches and treat two patches $(i,j)$ as opposed when their associated object-surface normals have a dot product below $-0.9$, allowing contacts from either one or two hands. For each opposed pair, we compare the distance between the hand-patch centroids $\mathbf c_{t,i}^{h}$ and $\mathbf c_{t,j}^{h}$ with the distance between their nearest object-surface points $\mathbf c_{t,i}^{o}$ and $\mathbf c_{t,j}^{o}$. When the hand grips the object with fingers on opposite sides, $\ell_{t,ij}^{\mathrm{obj}}$ and $\ell_{t,ij}^{\mathrm{hand}}$ should be equal. If not, the object’s anisotropic scale is corrected until they match.

\begin{equation}
\begin{aligned}
\ell_{t,ij}^{\mathrm{obj}}
&=\|\mathbf c_{t,i}^{o}-\mathbf c_{t,j}^{o}\|_2,
&
\ell_{t,ij}^{\mathrm{hand}}
&=\|\mathbf c_{t,i}^{h}-\mathbf c_{t,j}^{h}\|_2,\\
r_{t,ij}^{\mathrm{cal}}
&=\ell_{t,ij}^{\mathrm{obj}}
-\ell_{t,ij}^{\mathrm{hand}} .
\end{aligned}
\end{equation}

Let $\mathcal O_t$ denote the valid opposed pairs and $w_{t,ij}$ their
confidence. We define the caliper loss as
\begin{equation}
\mathcal L_{\mathrm{cal}}
=
10^3
\frac{
\displaystyle
\sum_{\substack{t\in\mathcal B_{\mathrm{cal}}\\(i,j)\in\mathcal O_t}}
w_{t,ij}\,
\rho_{c_{\mathrm{tac}}}\!\left(r_{t,ij}^{\mathrm{cal}}\right)
}{
\displaystyle
\sum_{\substack{t\in\mathcal B_{\mathrm{cal}}\\(i,j)\in\mathcal O_t}}
w_{t,ij}+\epsilon
}.
\label{eq:caliper-loss}
\end{equation}

\subsubsection{Contact-Spread Scale Floor}
\label{sec:spread}
Any two simultaneous contact patches, from the same or different hands, provide a lower bound on the object's diameter. Unlike the caliper term, this constraint does not require opposed surface normals. To reduce errors from uncertain wrist depth and finite patch size, we retain only the separation perpendicular to the camera ray and subtract both patch radii.
\begin{equation}
\begin{aligned}
\Delta\mathbf c_{t,ij}^{h}
&=\mathbf c_{t,j}^{h}-\mathbf c_{t,i}^{h},\\
\ell_{t,ij}^{\mathrm{floor}}
&=
\left\|
(\mathbf I_3-\mathbf d_{t,ij}\mathbf d_{t,ij}^{\!\top})
\Delta\mathbf c_{t,ij}^{h}
\right\|_2
-\varrho_{t,i}-\varrho_{t,j}.
\end{aligned}
\label{eq:spread}
\end{equation}

We estimate a robust scale floor from readings near the clip's $95$th percentile. Let $r$ be the ratio of this floor to the prior's longest extent. When $r\in[1.03,1.6]$, we initialize the corresponding log-scale correction with $\log r$ and prevent optimization from shrinking the object below the measured floor. Smaller ratios are uninformative, while larger ratios are rejected as likely outliers.

\subsection{Hand Objective}
\label{sec:hand-objective}

For each hand $h$, the hand objective is defined as follows.
\begin{equation}
\begin{split}
\mathcal L_{\mathrm{hand}}^h=
&\mathcal L_{\mathrm{reproj}}^h
+10\mathcal L_{\mathrm{depth}}^h
+\mathcal L_{\mathrm{prior}}^h
+5\mathcal L_{\mathrm{range}}^h
\\
&+\lambda_{\mathrm{temp}}\mathcal L_{\mathrm{temp}}^h
+100\mathcal L_{\mathrm{pen}}^h
\label{eq:hand-objective}
\end{split}
\end{equation}

The hand objective uses an EMA-normalized mean 2-D joint reprojection loss $\mathcal L_{\mathrm{reproj}}^h$ to anchor the hand in the image plane.
A wrist-depth loss $\mathcal   L_{\mathrm{depth}}^h$ penalizes the squared difference between the optimized wrist depth and the HaMeR estimate, interpolated when unavailable. The initialization prior $\mathcal L_{\mathrm{prior}}^h$ regularizes MANO pose and shape toward their stage-1 estimates, while $\mathcal L_{\mathrm{range}}^h$ penalizes pose parameters with magnitude above $2$ radians.

The temporal loss $\mathcal L_{\mathrm{temp}}^h$ discourages abrupt hand motion by penalizing second-order differences in translation, 6-D wrist rotation, and articulation. We set $\lambda_{\mathrm{temp}}=10$ for sequences recorded at 30\,fps and reduce it to $0.1$ for the temporally subsampled ARCTIC and HO3D sequences, where adjacent frames are farther apart.

The penetration loss $\mathcal L_{\mathrm{pen}}^h$ prevents implausible hand--object intersections. For each candidate hand vertex $\mathbf v_{t,q}^h$, we find its nearest object-surface point $\mathbf c_{t,q}^o$ and outward normal $\mathbf n_{t,q}^o$, and compute
$\delta_{t,q}^h=(\mathbf v_{t,q}^h-\mathbf c_{t,q}^o)^\top\mathbf n_{t,q}^o$.
Negative values indicate that the vertex lies inside the object. We activate the loss for vertices with more than $1\,\mathrm{cm}$ of penetration, apply a $5\,\mathrm{mm}$ residual offset, and cap the residual at $4\,\mathrm{cm}$ to limit extreme gradients. The nearest surface points and normals are detached during optimization, so this term updates the hand without moving the object.

\subsection{Implementation Details}
\label{sec:implementation}

We optimize Eq.~\eqref{eq:joint-objective} with Adam~\cite{adam} for 600 iterations using an initial learning rate of $3\times10^{-3}$ and cosine annealing~\cite{sgdr}. Each iteration samples 12 frames for the shared hand-object terms and up to eight additional frames from those on which the hand is detected. We freeze $\boldsymbol\alpha$ for the first $33\%$ of iterations so that pose converges before scale can absorb pose error. We render soft silhouettes with PyTorch3D~\cite{pytorch3d}, and contact-candidate vertices follow DexGraspNet~\cite{dexgraspnet}. Across 54 ARCTIC fits, joint refinement required a median of $50\,$s per clip, with a range of $42$--$111\,$s. PyTorch3D's nondeterministic CUDA atomics produce run-to-run mIoU differences of approximately $2\times10^{-5}$.

\section{Experiments}
\label{sec:experiments}
\begin{figure*}[h]
\centering
\includegraphics[
    width=0.8\textwidth,
    keepaspectratio
]{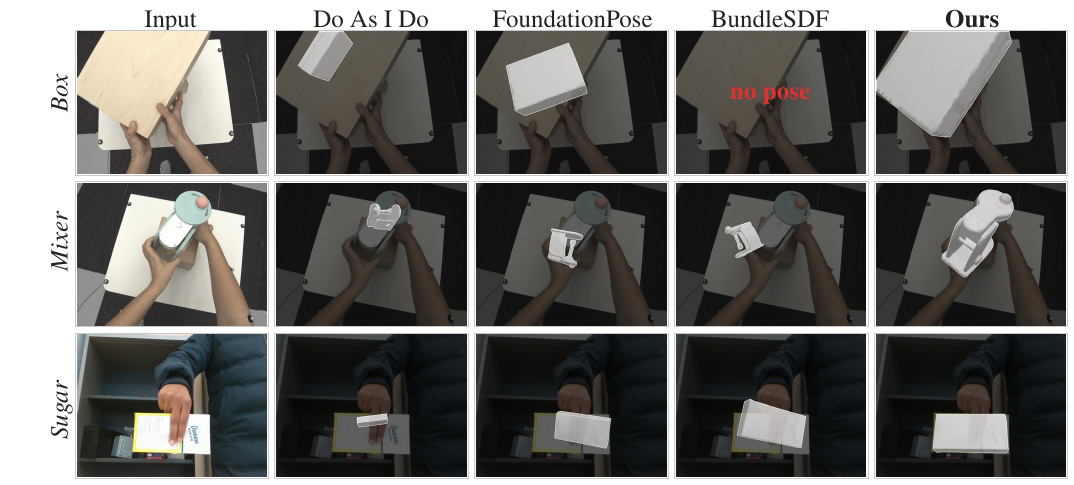}
\caption{Qualitative camera-space pose comparison. These overlays place the ground truth mesh at the predicted pose, isolating the pose produced by each method. BundleSDF fails to produce a pose for 48/150 frames on the box clip.}
\label{fig:pose-comparison}
\end{figure*}
We evaluate whether tactile contact can serve as an effective metric constraint for object pose and reconstruction. We assess camera-space tracking accuracy, analyze the contributions of individual loss groups, evaluate object geometry and scale, and demonstrate the complete pipeline using measured tactile data on in-the-wild OpenTouch clips.

\subsection{Experimental Setup}
\label{sec:exp-setup}

\noindent \textbf{Datasets and tactile observations.}
ARCTIC~\cite{arctic} contributes 55 egocentric bimanual clips spanning 11 articulated objects and seven subjects. It provides object and MANO ground truth but no sensor depth. HO3D~\cite{ho3d} contributes 45 single-hand clips covering nine YCB objects and provides object, hand, and RGB-D ground truth. We evaluate five clips per object and subsample each clip to at most 150 frames. Neither benchmark records touch, so we synthesize the map from ground truth object and hand pose. OpenTouch~\cite{opentouch} provides 30 in-the-wild clips with a measured $16\!\times\!16$ pressure glove and motion-capture hand skeleton, but no object-pose, object-geometry, or depth ground truth.

\noindent \textbf{Baselines and coverage.}
We compare TacHOI against FoundationPose~\cite{foundationpose}, BundleSDF~\cite{bundlesdf}, and ForeHOI~\cite{forehoi}. FoundationPose is provided the ground-truth CAD model. On HO3D, FoundationPose, Do As I Do, and BundleSDF use sensor depth, and on ARCTIC they use ViPE-estimated monocular depth. These comparisons are therefore informative but not strictly input-matched. We exclude HOLD~\cite{hold} and Magic HOI~\cite{magichoi} because their per-sequence runtime makes evaluation at our benchmark scale impractical. ViTaM~\cite{vitam} is not directly comparable because it relies on a learned distributed-tactile framework and targets dynamic interaction reconstruction rather than metric object pose and scale.

\noindent \textbf{Metrics.}
ADD-S$^{*}$ is the directed mean nearest-surface distance from the posed prediction to ground truth. The asterisk distinguishes this non-corresponding-mesh variant, which includes both reconstruction and pose error. We also report centroid error, along-ray error, and rotation error. Shape metrics use similarity ICP followed by bidirectional Chamfer distance (CD) and F-score at 5 and 10\,mm. Similarity alignment removes translation, rotation, and scale and therefore evaluates shape rather than metric reconstruction. Scale error is reported analogously as the
magnitude of the signed median. We compute a statistic per clip and report the median across all clips per dataset.

\subsection{Metric Camera-Space Tracking}
\label{sec:tracking-results}
\begin{table}[h]
\centering
\caption{Metric camera-space tracking without alignment or scale normalization.}
\label{tab:pose-main}
\scriptsize
\setlength{\tabcolsep}{3pt}
\resizebox{\columnwidth}{!}{%
\begin{tabular}{@{}lcccc@{}}
\toprule
Method / Metric Constraint &
\shortstack{ADD-S$^{*}$\\(cm)$\downarrow$} &
\shortstack{Centroid\\(cm)$\downarrow$} &
\shortstack{Along-Ray Error\\(cm)$\downarrow$} &
\shortstack{Rotation\\($^\circ$)$\downarrow$} \\
\midrule

\multicolumn{5}{l}{\emph{ARCTIC}} \\
Do As I Do track / ViPE
& 15.6 & 16.4 & 17.5 & 41.5 \\
FoundationPose / ViPE
& 35.2 & 39.2 & 35.0 & 70.4 \\
BundleSDF / ViPE
& 21.4 & 22.5 & 21.4 & 62.7 \\
\textbf{TacHOI / hand + contact}
& \textbf{3.6} & \textbf{5.6} & \textbf{1.8} & \textbf{40.1} \\
\midrule

\multicolumn{5}{l}{\emph{HO3D}} \\
Do As I Do track / sensor
& 3.6 & 5.3 & 3.4 & 58.2 \\
FoundationPose / sensor
& 7.4 & 7.6 & 4.8 & \textbf{43.1} \\
BundleSDF / sensor
& 3.3 & \textbf{2.8} & \textbf{0.6} & 51.7 \\
\textbf{TacHOI / hand + contact}
& \textbf{2.8} & 4.4 & \textbf{0.6} & 57.6 \\
\bottomrule
\end{tabular}%
}
\end{table}
Table~\ref{tab:pose-main} reports the primary result. On ARCTIC, which has no depth sensor, TacHOI reduces ADD-S$^{*}$ from 15.6\,cm for the closest baseline to 3.6\,cm and centroid error from 16.4 to 5.6\,cm. TacHOI reduces along-ray error from 17.5 to 1.8\,cm. On HO3D, TacHOI achieves the lowest ADD-S$^{*}$ at 2.8\,cm with full coverage, while BundleSDF obtains lower centroid and along-ray errors on the 77\% of frames for which it returns a pose. Rotation remains challenging on HO3D, even though FoundationPose benefits from the ground-truth CAD model. Overall, these results show that contact with a calibrated hand provides an effective metric constraint for monocular object tracking.

\subsection{Object Shape and Scale}
\label{sec:shape-results}

\begin{table}[h]
\caption{Shape after similarity ICP and relative metric scale error.}
\label{tab:shape}
\centering
\scriptsize
\setlength{\tabcolsep}{3pt}
\resizebox{\columnwidth}{!}{%
\begin{tabular}{@{}lcccc@{}}
\toprule
Method &
\shortstack{F@5\,mm\\(\%)$\uparrow$} &
\shortstack{F@10\,mm\\(\%)$\uparrow$} &
\shortstack{CD\\(cm)$\downarrow$} &
\shortstack{Scale Error\\(\%)$\downarrow$} \\
\midrule

\multicolumn{5}{l}{\emph{ARCTIC}} \\
BundleSDF
& 30.4 & 49.5 & 2.59 & 31.0 \\
ForeHOI
& 38.2 & 60.7 & 1.98 & 55.0 \\
SAM 3D prior
& 45.0 & 71.2 & 1.61 & 12.0 \\
\textbf{TacHOI}
& \textbf{47.3} & \textbf{71.5} & \textbf{1.49} & \textbf{4.0} \\
\midrule

\multicolumn{5}{l}{\emph{HO3D}} \\
BundleSDF
& 40.9 & 68.3 & 1.58 & 7.0 \\
ForeHOI
& 70.2 & 90.9 & 0.87 & 24.0 \\
SAM 3D prior
& 73.9 & 92.9 & 0.78 & 11.0 \\
\textbf{TacHOI}
& \textbf{77.4} & \textbf{94.2} & \textbf{0.76} & \textbf{5.0} \\
\bottomrule
\end{tabular}%
}
\end{table}

Table~\ref{tab:shape} evaluates canonical geometry after similarity alignment. TacHOI achieves the strongest scores on both datasets, but improvements over the SAM 3D prior are modest: F@5 increases by 2.3 points on ARCTIC and 3.5 points on HO3D. This is expected because TacHOI adjusts anisotropic extent rather than explicitly refining mesh shape, so most object geometry is inherited from SAM 3D Objects. Tactile observations also provide limited shape information in these sequences, since subjects typically maintain a similar grasp throughout each clip, leaving very little data for tactile-based refinement to work with. 

\subsection{Sources of Metric Accuracy}
\begin{table}[h]
\centering
\caption{Ablation of the loss groups.}
\label{tab:ablation}
\scriptsize
\setlength{\tabcolsep}{3pt}
\resizebox{\columnwidth}{!}{%
\begin{tabular}{@{}lcccccc@{}}
\toprule
&
\shortstack{Centroid\\(cm)$\downarrow$} &
\multicolumn{2}{c}{Along-Ray Error (cm)$\downarrow$} &
\multicolumn{2}{c}{Scale Error (\%)$\downarrow$} &
\shortstack{MPJPE\\(cm)$\downarrow$} \\
\cmidrule(lr){3-4}
\cmidrule(lr){5-6}
Configuration & & Median & Mean & Median & Mean & \\
\midrule

\multicolumn{7}{l}{\emph{ARCTIC} ($n=22$)} \\
Full objective
& 4.9 & \textbf{0.1} & 1.1 & 1.9 & \textbf{0.8} & 3.5 \\
w/o hand terms
& 4.8 & 0.7 & 1.6 & 1.2 & 1.8 & 6.4 \\
w/o image terms
& 5.9 & 1.1 & \textbf{0.3} & 3.6 & 0.8 & 3.5 \\
w/o contact + occl.
& 4.8 & 0.9 & 1.6 & 1.2 & 1.6 & \textbf{3.4} \\
w/o metric scale
& \textbf{4.7} & 1.6 & 2.2 & 1.6 & 3.5 & 3.5 \\
w/o regularizers
& 5.2 & 0.3 & 1.0 & \textbf{0.7} & 0.9 & 3.5 \\
\midrule

\multicolumn{7}{l}{\emph{HO3D} ($n=18$)} \\
Full objective
& 4.2 & 0.3 & \textbf{0.5} & 4.3 & 11.9 & \textbf{3.8} \\
w/o hand terms
& 4.7 & \textbf{0.1} & 0.8 & 4.6 & 12.1 & 7.4 \\
w/o image terms
& 6.7 & 2.2 & 1.7 & \textbf{1.0} & \textbf{7.6} & 4.3 \\
w/o contact + occl.
& 5.1 & 1.2 & 1.4 & 4.7 & 12.1 & 4.0 \\
w/o metric scale
& \textbf{3.9} & 0.2 & 1.0 & 8.2 & 13.1 & \textbf{3.8} \\
w/o regularizers
& 4.5 & 0.2 & 0.7 & 5.0 & 12.3 & \textbf{3.8} \\
\bottomrule
\end{tabular}%
}
\end{table}
\textbf{Grouped ablation.}
Table ~\ref{tab:ablation} shows a loss-group ablation on a balanced subset containing two clips per object (22 ARCTIC and 18 HO3D clips). Each row removes one group from the full objective. Every arm starts from the same stage-1 initialization on the same frames. Removing the image terms lowers scale error but nothing then anchors the object to the mask, so centroid error rises to 5.9 and 6.7 cm. Along-ray error grows to over 1 cm. Removing the contact terms leaves object placement nearly unchanged but raises along-ray error from 0.1 to 0.9 cm on ARCTIC and 0.3 to 1.2 cm on HO3D, since depth is no longer tied to the hand. Removing the metric-scale terms yields the lowest centroid error on both datasets but nearly doubles HO3D scale error (4.3 to 8.2\%), and removing the hand terms barely affects the object while doubling hand error (3.5 to 6.4 cm, 3.8 to 7.4 cm). The full objective is the only configuration that avoids a large loss in any column, and it gives the lowest along-ray error on ARCTIC, the quantity contact is meant to constrain.

\subsection{OpenTouch Evaluation}
\label{sec:real-results}
\begin{figure}[h]
\centering
\includegraphics[width=\columnwidth]{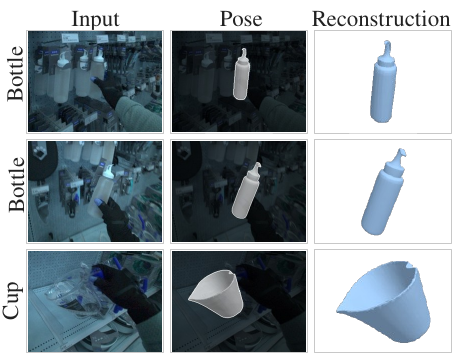}
\caption{Measured-tactile OpenTouch qualitative results on three objects in the grocery subset.}
\label{fig:opentouch-qual}
\end{figure}
OpenTouch dataset is particularly challenging due to high clutter and quick movements. TacHOI often struggles with these clips, but, on clips where SAM 3D can create a good enough mesh, and the hand can be tracked well, our method produces exceptional results. This experiment establishes end-to-end operation with measured pressure, but metric accuracy on real data requires independently measured object pose or dimensions. 

\section{Conclusion}
We presented TacHOI, a framework that jointly refines hand poses, object motion, and metric scale by combining monocular visual estimates with tactile contact. Our results show that contact with a calibrated hand provides a metric reference beyond image consistency: contact constrains object depth, while separated contacts constrain object extent. The resulting gains primarily concern metric placement and scale, highlighting the value of touch for grounding visually plausible reconstructions in physical dimensions.

% TacHOI bibliography, verified against primary proceedings, publisher,
% project, and arXiv records on 2026-09-15.
% Usage: place \input{references} where the reference list should appear.
% The keys are method-oriented so they remain easy to use in the manuscript.

% \input{sec/exp}

\bibliographystyle{IEEEtran}
%\bibliography{references}

\end{document}